\documentclass[lettersize,journal]{IEEEtran}
\usepackage{amsmath,amsfonts}
\usepackage{algorithm}
\usepackage{algpseudocode}
\usepackage{array}
\usepackage{textcomp}
\usepackage{stfloats}
\usepackage{url}

\usepackage{verbatim}
\usepackage{subcaption}
\usepackage{makecell}
\usepackage{caption}
\usepackage{graphicx}
\usepackage{cite}
\usepackage{placeins}
\usepackage{overpic}
\usepackage{hyperref}

\newif\ifuseRevisions
\useRevisionsfalse  % Change to \useRevisionsfalse to disable blue text
\newcommand{\rev}[1]{\ifuseRevisions\textcolor{blue}{#1}\else #1\fi}

\usepackage[font=small,labelfont=bf]{caption}

\begin{document}

\title{Miniature multihole airflow sensor for lightweight aircraft over wide speed and angular range}

\author{Lukas Stuber$^*$, Simon Jeger$^*$, Raphael Zufferey, Dario Floreano,~\IEEEmembership{Fellow,~IEEE,}
        % <-this % stops a space
\thanks{$^*$These authors contributed equally. All authors are with the Laboratory of Intelligent Systems, Ecole Polytechnique Federale de Lausanne (EPFL), CH1015 Lausanne, Switzerland.}}% <-this % stops a space


\maketitle

\begin{abstract}
%Describe your work briefly: what is the problem, what others did, what you contributed, experimental results/comparisons, and conclusion.
An aircraft's airspeed, angle of attack, and angle of side slip are crucial to its safety, especially when flying close to the stall regime. Various solutions exist, including pitot tubes, angular vanes, and multihole pressure probes. However, current sensors are either too heavy ($> $30 g) or require large airspeeds ($> $20~m/s), making them unsuitable for small uncrewed aerial vehicles.
We propose a novel multihole pressure probe, integrating sensing electronics in a single-component structure, resulting in a mechanically robust and lightweight sensor (9 g), \rev{which we released to the public domain}.
Since there is no consensus on two critical design parameters, tip shape (conical vs spherical) and hole spacing (distance between holes), we provide a study on measurement accuracy and noise generation using wind tunnel experiments.
The sensor is calibrated using a multivariate polynomial regression model over an airspeed range of 3-27~m/s and an angle of attack/sideslip range of $\pm$35$^\circ$, achieving a mean absolute error of 0.44~m/s and 0.16$^\circ$. 
Finally, we validated the sensor in outdoor flights near the stall regime. Our probe enabled accurate estimations of airspeed, angle of attack and sideslip during different acrobatic manoeuvres.
Due to its size and weight, this sensor will enable safe flight for lightweight, uncrewed aerial vehicles flying at low speeds close to the stall regime.
\end{abstract}


\section{Introduction}
% UAVs are important
\IEEEPARstart{W}{winged} uncrewed aerial vehicles (UAVs) are transforming industries like agriculture, environmental monitoring, surveillance, and disaster management with their ability to access remote areas due to their large flight endurance \cite{floreano2015science}.

% Stalling is a problem.
However, stalling poses a critical threat to these UAVs. It results in a sudden loss of lift that can cause rapid altitude drops and uncontrollable spins, often leading to crashes \cite{brigenair,northwest,lin2020failure}. 
Small UAVs are particularly prone to stalling since wind gusts affect them more strongly than heavier, larger aeroplanes. 
To mitigate these risks, accurately estimating the three primary airflow components—airspeed, angle of attack (AoA), and angle of sideslip (AoS)—is essential. This enables early stall detection, reducing the likelihood of dangerous situations and expanding the UAV's operational flight envelope to allow for dynamic and acrobatic manoeuvres.

% What general methods exist to fix that problem
Methods for synthetic airflow estimation use inertial sensors and GNSS measurements to predict wind and large-scale turbulence, enabling the estimation of airspeed, AoA, and AoS. However, they typically rely on accurate aerodynamic models \cite{sharma2024synthetic} or assume steady wind conditions \cite{johansen2015estimation}, which limits their effectiveness for stall detection in real-world flight scenarios where wind and turbulence are variable.
The pitot tube remains the most common sensor for measuring airspeed on UAVs due to its simplicity and reliability. It estimates airspeed by measuring the difference between the air pressure at the tip (stagnation pressure) and along its side (static pressure) \cite{small_UAV_standard_sensors}. However, as the pitot tube only captures airspeed along the UAV's primary flight direction, it is typically used alongside two angular vane sensors — one for AoA and one for AoS measurements. These vane sensors consist of thin blades that rotate freely around an off-centre axis, aligning themselves with the airflow \cite{Distributed_pressure_sensing}.
This triple sensor setup is complex, heavy, and requires fuselage or wing mounts, making it unsuitable for small UAVs.

\begin{figure}
    \centering
    \includegraphics[width=1\columnwidth]{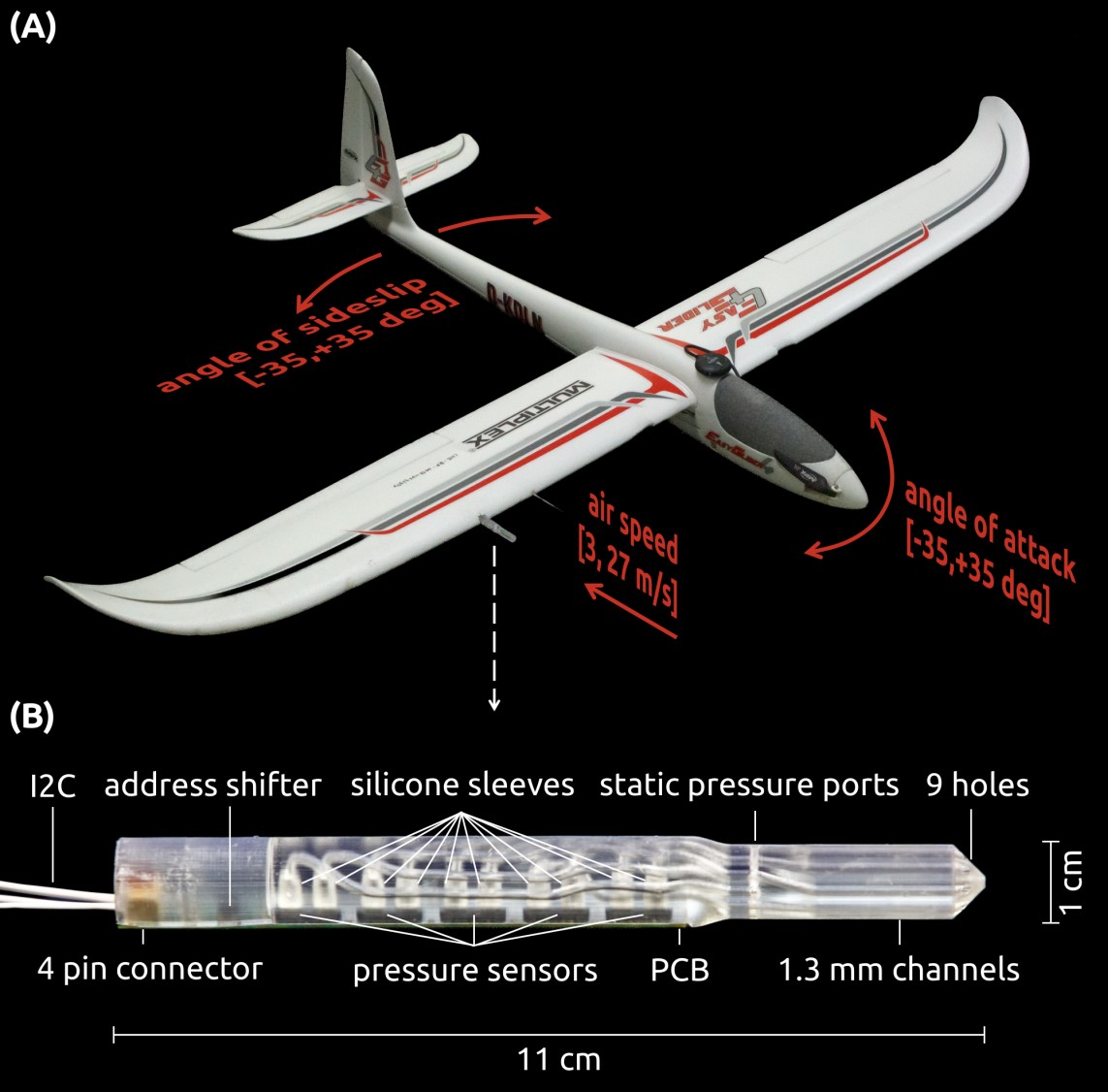}
    \caption{\textbf{Miniature integrated 3D airflow sensor: (A)} The fixed-wing UAV used for validation flights with the multihole pressure probe mounted on the right wing. \textbf{(B)} 
    The multihole pressure probe consists of a 3D-printed structure and sensor-PCB, joined by silicone sleeves that ensure an airtight seal. The integrated design eliminates traditional tubing between the structure and the sensors, significantly reducing weight and complexity.}
    \label{fig:overview}
\end{figure}

% Multihole pressure probes are a good way to fix that problem
Multihole pressure probes (MPPs) offer a robust solution for measuring airspeed, angle of attack (AoA), and angle of sideslip (AoS) in a single, compact sensor \cite{Johansen_compressible_calibration, johansen2015estimation, Al-Ghussain_aircraft_motion_bias}. Like pitot tubes, MPPs work by comparing stagnation and static pressures, but incorporate multiple pressure-sensing ports, allowing them to capture a pressure distribution which is used to estimate AoA and AoS. This design provides precise, reliable measurements without the need for moving parts, making MPPs a lightweight and efficient alternative to the traditional three-sensor setup.

\begin{figure}[t]
    \centering
    \begin{overpic}[width=1\columnwidth]{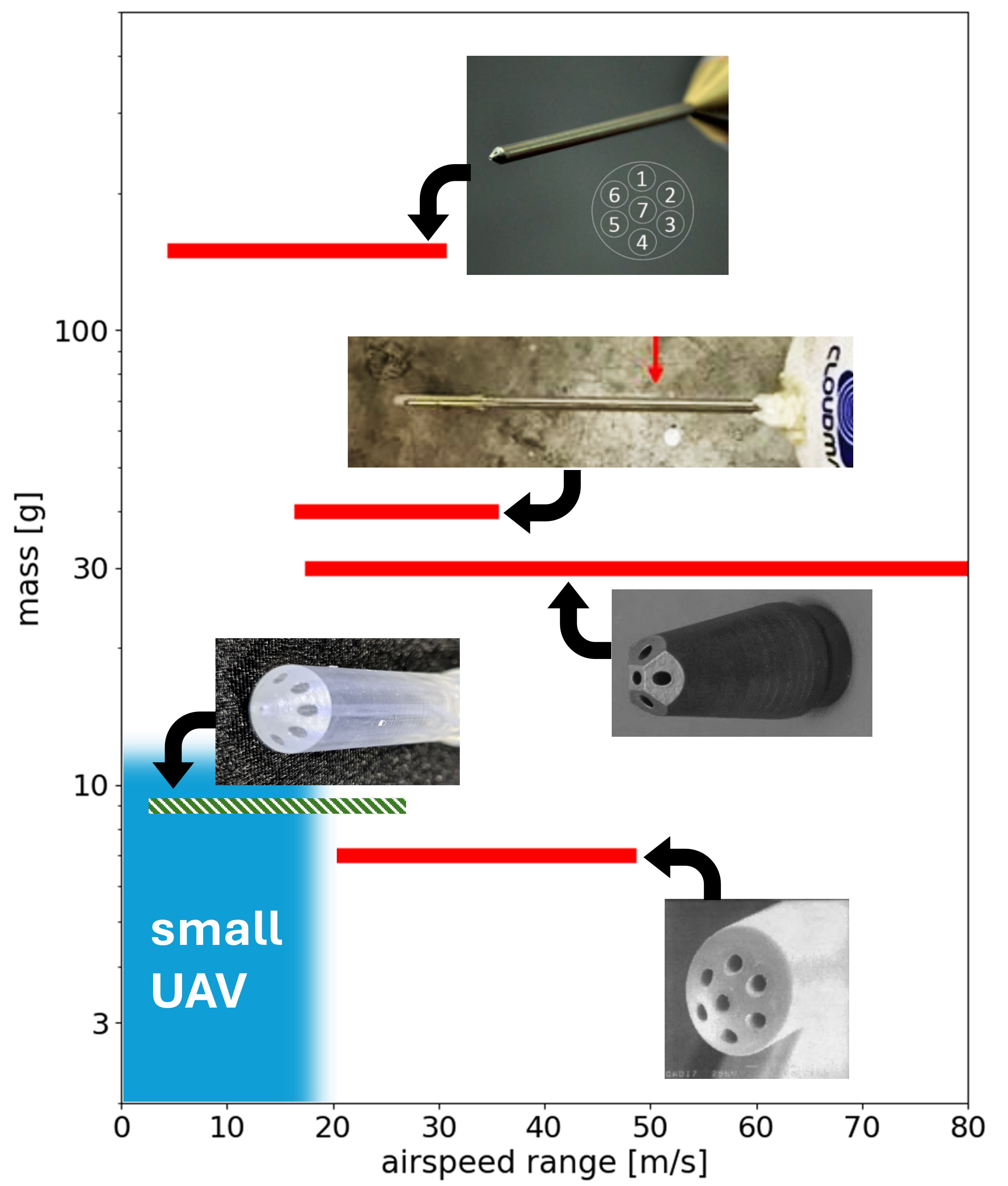}
        \put(62,85){\cite{Li_seven_hole_probe}}
        \put(72,65){\cite{Witte_development_uav}}
        \put(73.5,44){\cite{Hall_oxford_probe}}
        \put(72,16){\cite{Rediniotis_Tiny_probe}}
        \put(40,40){[ours]}
    \end{overpic}
    \caption{\textbf{Comparison of multihole pressure probes on mass and measurement range:} Existing sensors (red, \rev{full line}) are unsuitable for small uncrewed aerial vehicles (UAVs) due to their weight or airspeed range. Our approach (green, \rev{hatched line}) satisfies those requirements.}
    \label{fig:state_of_the_art}
\end{figure}
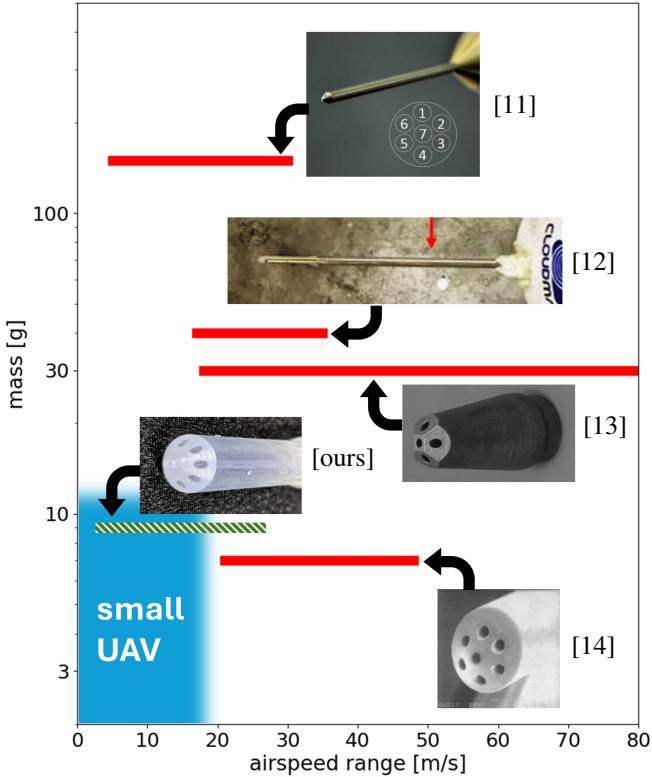

% But existing multihole pressure probes are too fat or require too much speed
However, existing lightweight MPPs require high airspeeds ($>$20~m/s) for reliable operation (Fig.~\ref{fig:state_of_the_art}). 
This prohibits the use of MPPs on lightweight UAVs, which typically fly at speeds below 15~m/s but cannot carry heavy sensory equipment. In this work, we propose a novel sensor (Fig.~\ref{fig:overview}) that fulfils both requirements: lightweight design (only 9~g) and functionality at low airspeeds (3 - 27~m/s). The proposed design is single-component which removes the need for any tubing, mounting, large channels in the wing and complex wing integration while minimising weight. \rev{To facilitate further development we release our design to the public domain.}

% There are some parameters that we need to explore
MPP are characterised by two design parameters: the tip shape, which affects airflow attachment at high angles, and the hole spacing, which determines the probe’s diameter. 
Existing MPP designs predominantly use either spherical tips \cite{crawford2013influence, Brown_nose_of_aircraft} or, more commonly, conical tips \cite{Hall_oxford_probe, Witte_development_uav, Li_seven_hole_probe}.
Although the literature is inconclusive, we expect larger hole spacing to result in higher measurement resolution at the cost of a greater diameter and, therefore, higher weight.
This study investigates the impact of various tip shapes and hole spacing on measurement resolution and noise generation in airflow regimes suitable for small UAVs ($<$15~m/s). The optimal design is characterised in a closed wind tunnel using multivariate polynomial regression and validated in outdoor flight tests.

\section{Hardware design} \label{sec:hardware_design}
MPPs are composed of two main components: a 3D-printed multihole structure containing internal airflow channels and a set of pressure sensors mounted on a printed circuit board (PCB). In conventional MPP designs, these components are mounted separately, with flexible tubing connecting each channel in the multihole structure to a corresponding pressure sensor on the PCB. Typically, the number of holes aligns with the number of sensors, with each sensor measuring absolute pressure from a single channel \cite{Li_seven_hole_probe, Hall_oxford_probe, Sankaralingam_low_cost_five_hole}.

Our proposed design differs in two key ways: First, we integrate the PCB directly within the multihole structure, eliminating the need for separate tubing. This integration not only simplifies the assembly but also reduces the overall weight and system complexity compared to conventional MPPs. Second, rather than configuring each sensor to measure absolute pressure from a single hole, our design measures the pressure differential between paired holes. This method effectively halves the number of required sensors, further reducing weight and enhancing efficiency \cite{Alaoui_Sosse_profiles_and_turbulence}.

\begin{figure}[h]
    \centering
    \includegraphics[width=1\columnwidth]{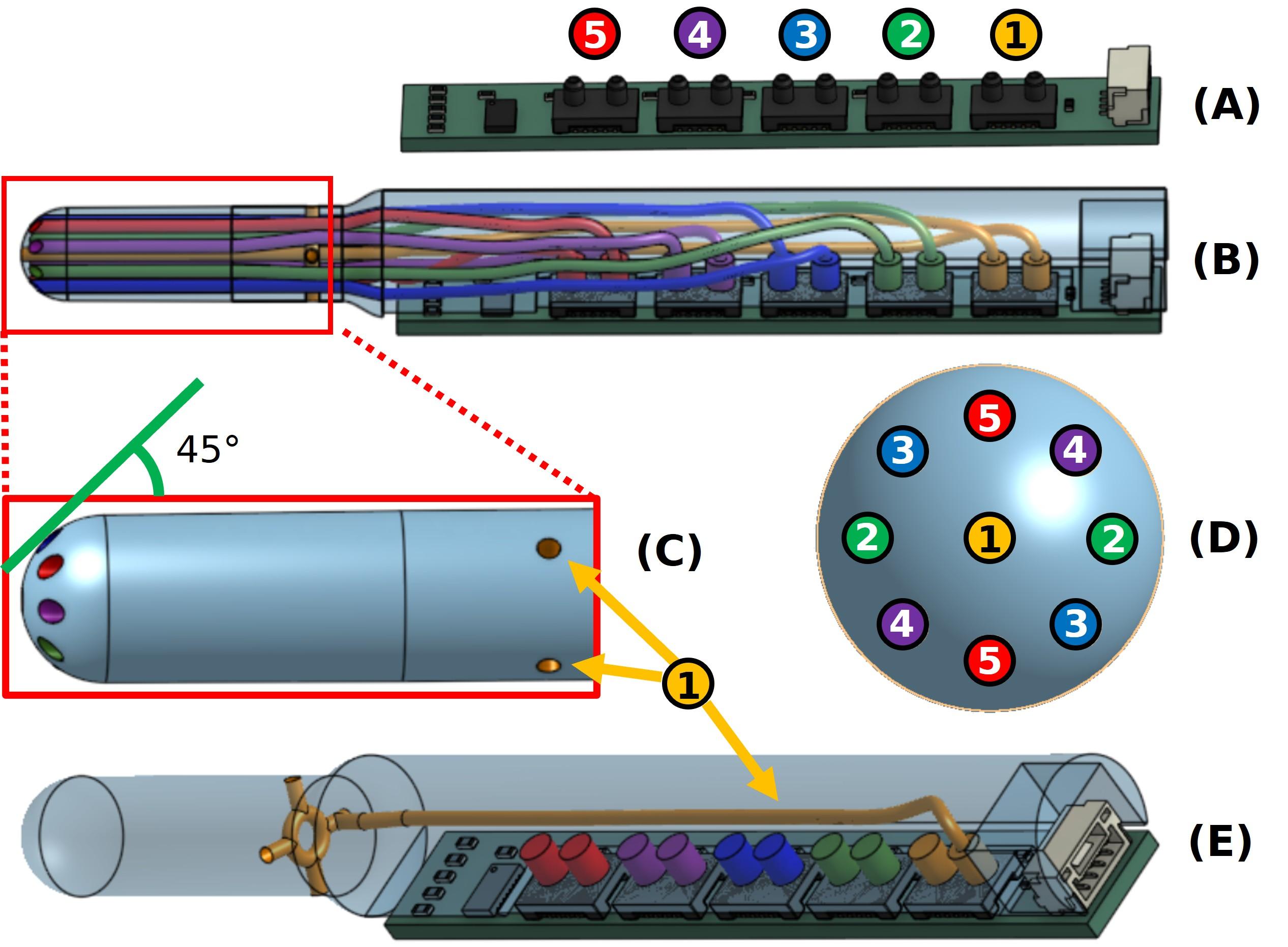}
    \caption{\textbf{Design of the multihole pressure probe:} 
    \textbf{(A)} Printed circuit board (PCB) with five differential pressure sensors (1-5). Each sensor is connected to two holes, measuring differential pressure. Data is transmitted via a connector in the back (right side).
    \textbf{(B)} Complete probe with colour-coded internal pathways.
    \textbf{(C)} Side view, showing the static holes at the side (yellow). Front holes are placed at a 45$^\circ$ angle.
    \textbf{(D)} Front view, displaying pairs of holes (e.g., left-right or up-down) that are connected to the same differential pressure sensor.
    \textbf{(E)} Isometric view, showing only the static hole for enhanced readability.}
    \label{fig:cad}
\end{figure}

\subsection{Probe description} \label{sec:probe_description}
The MPP is equipped with a PCB featuring five SDP31 differential pressure sensors, each capable of measuring pressure differences up to $\pm$500 Pa (Fig.\ref{fig:cad}A). The PCB is integrated directly within the 3D-printed multihole structure, with silicone sleeves providing airtight seals between the sensors and the internal channels of the multihole structure (Fig.~\ref{fig:cad}B).
This structure features 13 holes in total. The first hole is centrally located at the tip, with eight additional holes arranged in a circular pattern around it. These peripheral holes are angled at 45$^\circ$ relative to the probe’s main axis to enhance angular sensitivity \cite{Brown_nose_of_aircraft} (Fig.~\ref{fig:cad}C), and each adjacent pair of holes is connected to one differential pressure sensor (Fig.~\ref{fig:cad}D). The four remaining holes are positioned further down the probe, perpendicular to the airflow, to capture static pressure in a single channel (Fig.~\ref{fig:cad}C,E).

\subsection{Design parameter optimisation} \label{sec:parameter_optimization}
\begin{figure}[t]
    \centering
    \includegraphics[width=1\columnwidth]{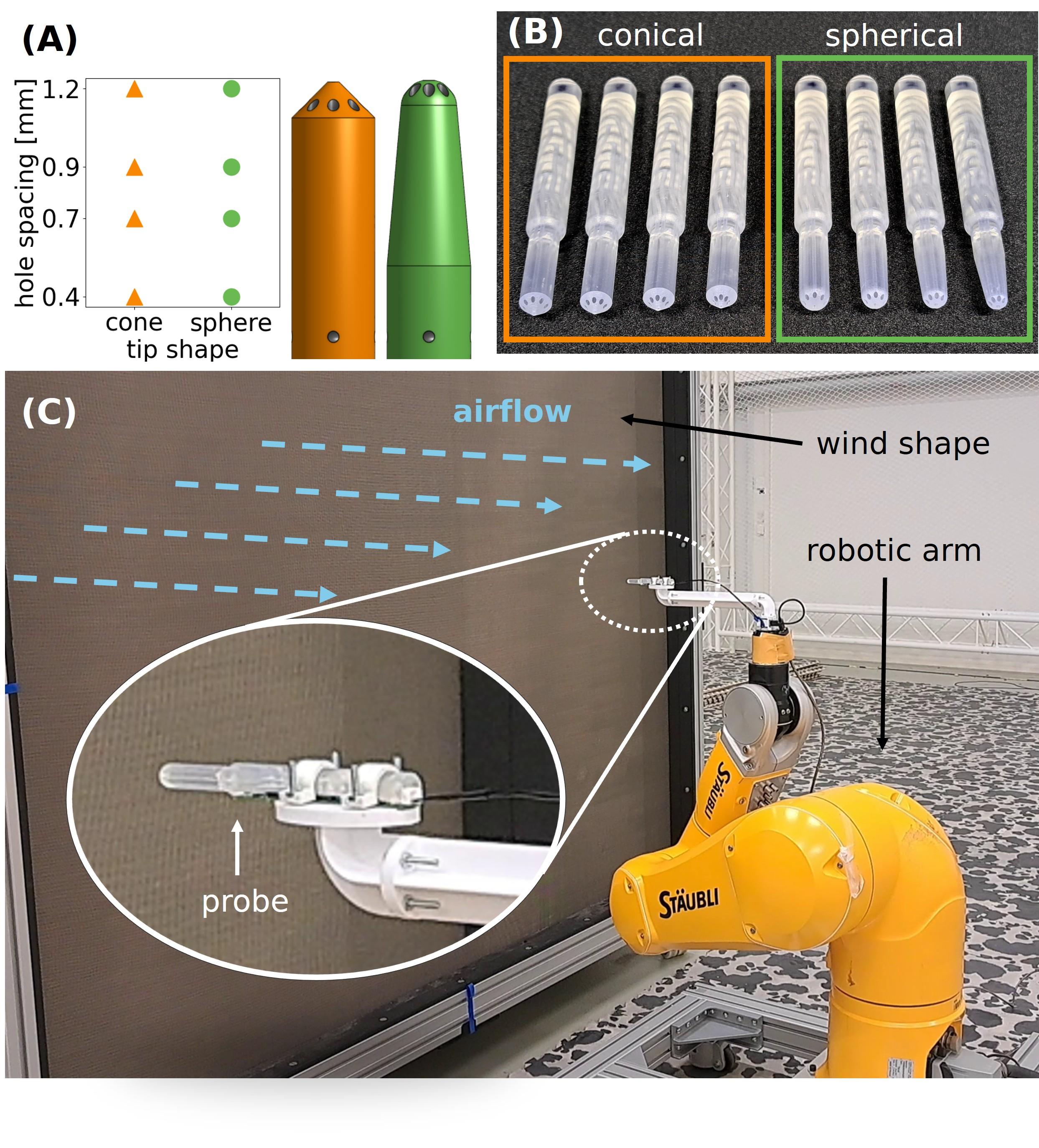}
    \caption{\textbf{Design parameter optimisation:}
    \textbf{(A)} Four different hole spacings between 0.4~mm and 1.2~mm are explored with cone and sphere tip shapes, resulting in eight different designs.
    \textbf{(B)} All probes were manufactured through Stereolithography resin printing.
    \textbf{(C)} A robotic arm was used to rotate the probes at different angles of attack and sideslip against an open wind tunnel (brown panel, Windshape Ltd)}
    \label{fig:hardware}
\end{figure}

To determine the optimal shape of the probe, eight designs were produced, investigating the influence of the probe's tip shape (cone, sphere) and hole spacing (0.4, 0.7, 0.9, 1.2~mm) on measurement resolution and noise generation (Fig.\ref{fig:hardware}A,B). The range of hole spacings was constrained by the minimal tolerances of the manufacturing process and the diameter of the tube. These designs were characterised in constant airflow from an open wind tunnel at 3, 6, 9, and 12~m/s. For each design and airspeed, a 6 DOF robotic arm positioned the MPPs in a sequence of 81 AoA and AoS configurations between $\pm$70$^\circ$ (Fig.~\ref{fig:hardware}C). Measurements were collected at 50~Hz for 2~s per airflow configuration (airspeed, AoA, AoS), resulting in a total of 259'200 data points.

\begin{figure}[h]
    \includegraphics[width=1\columnwidth]{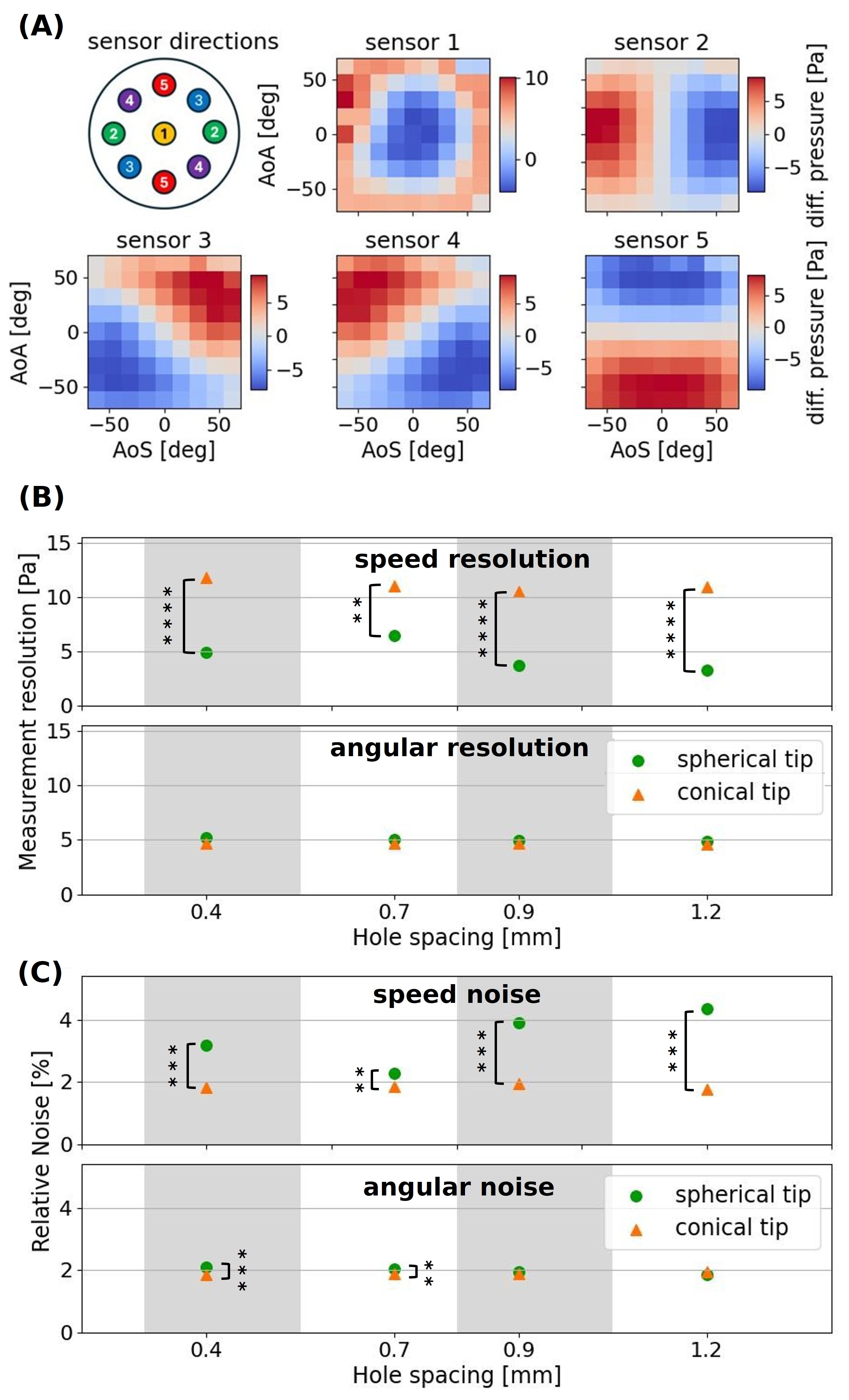}
    \caption{\textbf{Results of the design parameter optimisation: (A)} Differential pressures at 3~m/s airspeed, averaged over time (1.4~s / 70 samples), are plotted as a function of AoA and AoS. Each of the five pressure sensors records the pressure difference between a pair of ports. The distribution is aligned with the direction of the hole pair of the corresponding sensor, enabling estimation of AoA and AoS. \textbf{(B)} \rev{Comparison of all tested hardware configurations with respect to measurement resolution across velocity and angular ranges. Each point or triangle represents a distinct hardware design, with its performance compressed into a single metric according to the procedures outlined in Table~\ref{tab:compare_res_noise}. Statistically significant differences, as determined by independent t-tests, are marked with asterisks. The results show that conical tips yield significantly higher velocity measurement resolution compared to spherical tips, while hole spacing has no significant impact.
    \textbf{(C)} Comparison of hardware designs in terms of noise generation. Conical tips produce significantly lower speed noise compared to spherical tips.}}
    \label{fig:shape_evaluation}
\end{figure}

Sensor 1 (static pressure - tip centre pressure) shows a pressure difference that varies radially with AoA and AoS, reaching its minimum when the probe is aligned parallel to the airflow (AoA = AoS = 0), as expected. In this alignment, the channel functions similarly to a pitot tube, capturing the airflow component parallel to the probe’s axis \cite{pitot_tube} \rev{(See Fig.~\ref{fig:shape_evaluation}A)}.

For sensors 2-5, the pressure differences vary according to both AoA and AoS, showing gradients that align with the orientation of each sensor’s hole pair. For example, sensor 2 measures the pressure difference between the right and left holes, producing a pressure gradient primarily aligned along the horizontal axis. Similarly, sensor 3, which measures the difference between the upper-right and lower-left holes, shows a dominant gradient along the first diagonal, aligned with its hole pair orientation.

We compare the 8 designs in measurement resolution and noise generation (see Fig. \ref{fig:shape_evaluation}B). \rev{In order to maximise the sensors sensitivity to changes in airspeed and angles, the preferred design has a high measurement resolution and low noise (Tab. \ref{tab:compare_res_noise}).} Both metrics are evaluated on the data of pressure sensor 1 for the airspeed and on the data of sensors 2,3,4,5 for the angular measurements. The measurement matrix has five dimensions: 8 hardware designs x 4 speeds x 5 pressure sensors x 81 angles x 70 samples. To calculate the four comparative metrics, the dimension of the data matrix is reduced using the operations shown in table \ref{tab:compare_res_noise}. 

\begin{table}[h]
    \centering
    \begin{tabular}{c|c|c|c|c}
    \hline
        & \makecell{time \\ 70 samples} & 81 angles & 4 sensors & 4 speeds \\
        \hline
        angular resolution & avg & \textbf{std} & avg & avg \\
        angular noise      & \textbf{std} & avg & avg & avg \\
        \hline
        airspeed resolution & avg & avg & - & \textbf{std} \\
        airspeed noise      & \textbf{std} & avg & - & avg \\
        \hline
    \end{tabular}
    \caption{The operation used to reduce each dimension for each comparison metric. The comparison of angular measurements use the data from the pressure sensors 2,3,4,5, while the speed comparison only uses sensor 1. For each metric the defining dimension is reduced using the standard deviation while all other dimensions are reduced with the average. The angular resolution is characterised by the signal's change with angle, the speed resolution by the change with airspeed, and the noises by the change with time. \rev{Accordingly, the preferred sensor shows a high variance over speeds and angles and low variance over time.}}
    \label{tab:compare_res_noise}
\end{table}

\rev{The results show that cone-shaped tips provide significantly higher measurement resolution and lower noise levels compared to spherical tips, making them the preferred design choice. In contrast, hole spacing was found to have no significant impact on sensor performance. Additionally, the probe’s weight is unaffected by either tip shape or hole spacing, as its diameter is primarily determined by fabrication constraints; the minimum printable hole diameter of 1.3~mm and the required wall thickness of 0.5~mm (Fig. \ref{fig:cad}B,E). Larger spacings between holes simplify the removal of residual liquid resin after SLA printing, thereby improving production reliability.} Based on these findings, the remainder of this article will focus on a probe configuration featuring a cone-shaped tip and a hole spacing of 1.2~mm

\subsection{Manufacturing}
Both Fused Deposition Modelling (FDM) and Stereolithography (SLA) resin printing were explored. Today, only SLA printing provides the precision necessary for creating thin, airtight channels. The SLA-printed piece is formed in a resin bath, filling all cavities in the process. Once complete, those cavities require draining, constraining the minimal channel diameter in our case to 1.3~mm due to capillary effects. Silicone sleeves are used to ensure an airtight connection between the resin-printed multihole structure and the pressure sensors (Fig. \ref{fig:overview}B). \rev{Our CAD design can be found here} \cite{zenodo2024}.

\begin{figure}[h!]
    \centering
    \includegraphics[width=0.9\columnwidth]{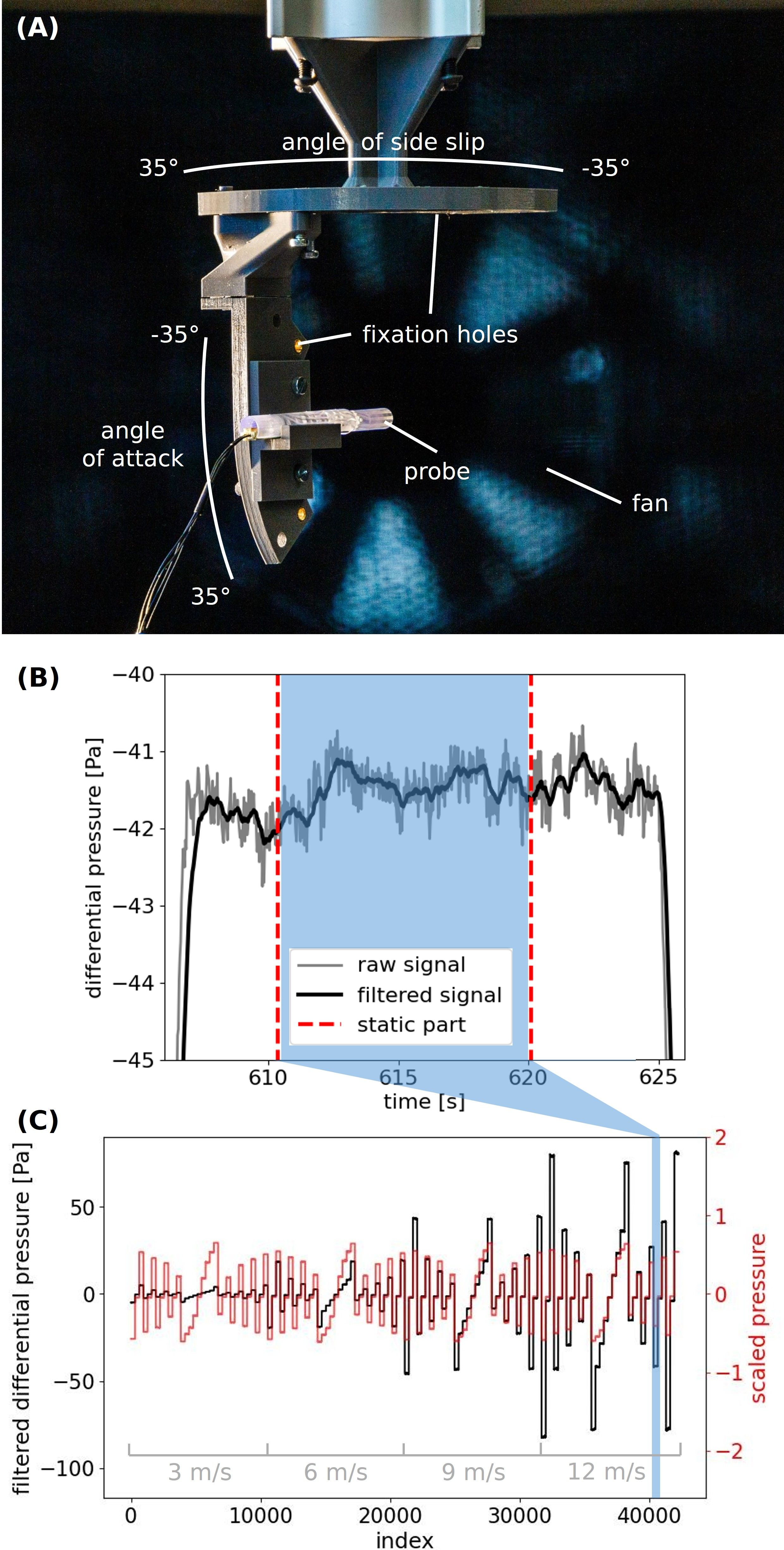}
    \caption{\textbf{Preprocessing:} 
     \textbf{(A)} A closed wind tunnel was used to gather differential pressure measurements at different speeds (3 - 27~m/s), AoA and AoS ($-35^\circ$ - $35^\circ$). The probe was mounted on a 3D-printed holder that can be oriented at five angles of attack (AoA) and five angles of sideslip (AoS) between $\pm$35$^\circ$ using fixation holes.
    \textbf{(B)} Filtering and slicing the measurements for calibration of pressure sensor 2 at 12~m/s speed, $-17.5^\circ$ AoA and $-17.5^\circ$ AoS. The raw measurement (grey) undergoes first-order Butterworth low-pass filtering with a cut-off frequency of 10 Hz (black). The static portion of the signal (middle 70\%, red) is used for calibration.
    \textbf{(C)} The filtered pressure differences of pressure sensor 2 over the first four air speeds (3, 6, 9, 12~m/s, at all AoA and AoS) (black). When scaled by the factor q (Eq.~\ref{eq:q_factor}) (red), the measurement is decoupled from the airspeed, simplifying the calibration process.}
    \label{fig:preprocessing}
\end{figure}
\section{Calibration} \label{sec:calibration}
MPPs measure pressure gradients from which airspeed, AoA, and AoS can be estimated. This process requires preprocessing to isolate the influence of airspeed from the angular components, followed by fitting a model to the processed data. The data was collected in a closed wind tunnel, where airspeeds ranged up to \rev{27~m/s}, across 17 different AoA and AoS configurations (Fig.~\ref{fig:preprocessing}A) each at nine different speeds (3, 6, 9, 12, 15, 18, 21, 24, 27~m/s). For each combination (Fig.~\ref{fig:preprocessing}B), the five differential pressure sensors recorded data at 33~Hz for 20~s, creating a total of 504900 measurements.

%However, this mapping is not linear; a calibration step is required. We use a multivariate polynomial regression model \cite{sinha2013multivariate} which is a common method for MPP calibration. The regression model takes the measurements from the five differential pressure sensors as input and produces one of the airflow components (speed, AoA or AoS) as output. Accordingly, three models, one for each component, are used. For accurate airflow estimation the model must be fitted with data ranging over the entirety of the airflow range that is expected during flights with small UAV. Unfortunately, the data set created for the hardware design choice only includes speeds up to 12~m/s due to the limited range of the open wind tunnel. A new data set with a speed range suitable for small UAV (0-30~m/s) was thus gathered in a closed wind tunnel capable of producing speeds up to 30~m/s (Fig.~\ref{fig:preprocessing}A,B). Measurements were done at 17 different AoA and AoS (Fig.~\ref{fig:calibration}A) each at nine different speeds (3, 6, 9, 12, 15, 18, 21, 24, 27~m/s), for a total of 153 combinations. For each combination, the five pressure differences were measured at 33~Hz for 20~s to generate the calibration data set.

\begin{figure*}[t]
     \centering
     \includegraphics[width=1\textwidth]{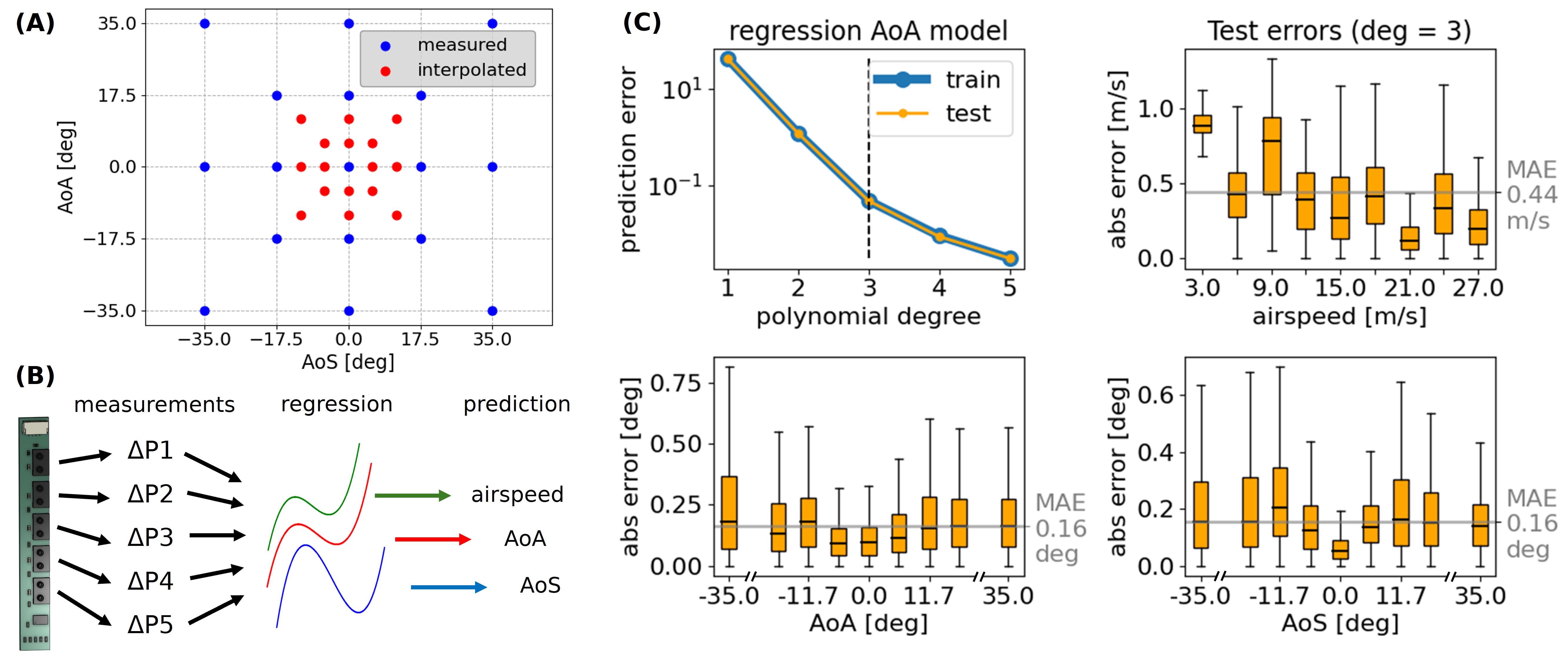}
     \caption{\textbf{Calibration method:}
     \textbf{(A)} The regression dataset includes 17 measured points (blue) obtained in a closed wind tunnel across nine different speeds ranging from 3~m/s to 27~m/s. To enhance numerical stability, additional linearly interpolated points (red) are added near the zero-angle regime.
     \textbf{(B)} \rev{Three separate polynomial regression models were fitted to predict airspeed, angle of attack, and angle of sideslip. Prior to model fitting, the dataset was randomly shuffled to ensure robustness. The exact polynomials can be found in our data repository \cite{zenodo2024}}.
     \textbf{(C)} \rev{The prediction error of the multivariate polynomial regression model, fitted to predict the angle of attack, is plotted against the polynomial degree, with a dashed line marking the elbow at degree three. The angle of attack model serves as a representative example for the airspeed and angle of sideslip models, which exhibit similar error characteristics. These three models are subsequently applied to the test subset for inference, showing a mean error of $0.44$~m/s and $0.16^\circ$.}}
     \label{fig:calibration}
\end{figure*}

\subsection{Preprocessing}
When the airflow is not parallel to the probe, both high airspeed and large AoA or AoS result in high differential pressure readings. To disentangle these two effects, the airspeed-related component of the signal is first calculated and used to scale all subsequent measurements. This approach is well-established for MPPs utilising absolute pressure sensors \cite{Witte_development_uav, Rediniotis_Tiny_probe, Hall_oxford_probe, Sankaralingam_low_cost_five_hole}. However, since our MPP relies on differential pressure sensors, this necessitates a novel preprocessing technique to effectively separate and handle the airspeed and angle-related components of the data.
We use the geometric mean of the pressure differences $P_i$ to calculate the scaling factor $q$ (Eq.~\ref{eq:q_factor}).

\begin{align}\label{eq:q_factor}
    q = \sqrt{ \Delta P_1^2 + \Delta P_2^2 + \Delta P_3^2 + \Delta P_4^2 + \Delta P_5^2}
\end{align}

Dividing the pressure differences by the scaling factor removes the influence of airspeed (Fig.~\ref{fig:preprocessing}C), allowing both components to be treated separately in the model fitting.

% \begin{align*}
%     X_{angle} &= [ \frac{\Delta P_1}{q}, \frac{\Delta P_2}{q}, \frac{\Delta P_3}{q}, \frac{\Delta P_4}{q}, \frac{\Delta P_5}{q},\\
%     &\hspace{6mm}\Delta P_1, \Delta P_2, \Delta P_3, \Delta P_4, \Delta P_5 ]^T
% \end{align*}
% while the scaling term $q$ together with the original pressure differences form the calibration data set for the speed $X_{speed}$.
% \begin{align*}
%     X_{speed} &= \left[ \Delta P_1, \Delta P_2, \Delta P_3, \Delta P_4, \Delta P_5, q \right]^T
% \end{align*}

\subsection{Model Fitting}
Using the dataset collected in the closed wind tunnel, we calibrate the sensor through multivariate polynomial regression \cite{sinha2013multivariate}. This method provides robust function approximation while maintaining a relatively simple model that generalises well to unforeseen scenarios. The data exhibit linear behaviour at small angles of attack (AoA) and sideslip (AoS), which we reinforce by generating artificial data through linear interpolation (Fig.~\ref{fig:calibration}A) before fitting the multivariate polynomial regression model (Fig.~\ref{fig:calibration}B). \rev{The dataset is first shuffled and subsequently divided into 70\% training and 30\% testing subsets.} Evaluation of the polynomial degree shows the best compromise of prediction accuracy and model complexity at degree three based on the elbow method \cite{marutho2018determination} (Fig.~\ref{fig:calibration}C). \rev{When evaluated on the test subset} the model achieved a mean absolute error (MAE) of 0.163$^\circ$ and 0.156$^\circ$ for the AoA and AoS estimation respectively and 0.44~m/s for air speed estimation (Fig.~\ref{fig:calibration}C). \rev{The speed error is highest at 3~m/s which marks the lower limit of our sensor's performance. The angular estimation is best around the zero-angle and shows no further deterioration beyond $11.7^\circ$ (Fig.~\ref{fig:calibration}C).
% \cite{Hall_oxford_probe} angle RMSE errors below 2° and speed errors below 4\% of speed\\
% \cite{Rediniotis_Tiny_probe} angle MAE errors of 0.28° and speed errors of 0.35\% \\
% \cite{Witte_development_uav} angle RMSE error under 0.13° and under 0.09m/s at 17m/s so under 0.53\%. \\
% We have between 0.56\% and 29.8\%
}

\begin{figure*}
    \centering
    \includegraphics[width=0.9\textwidth]{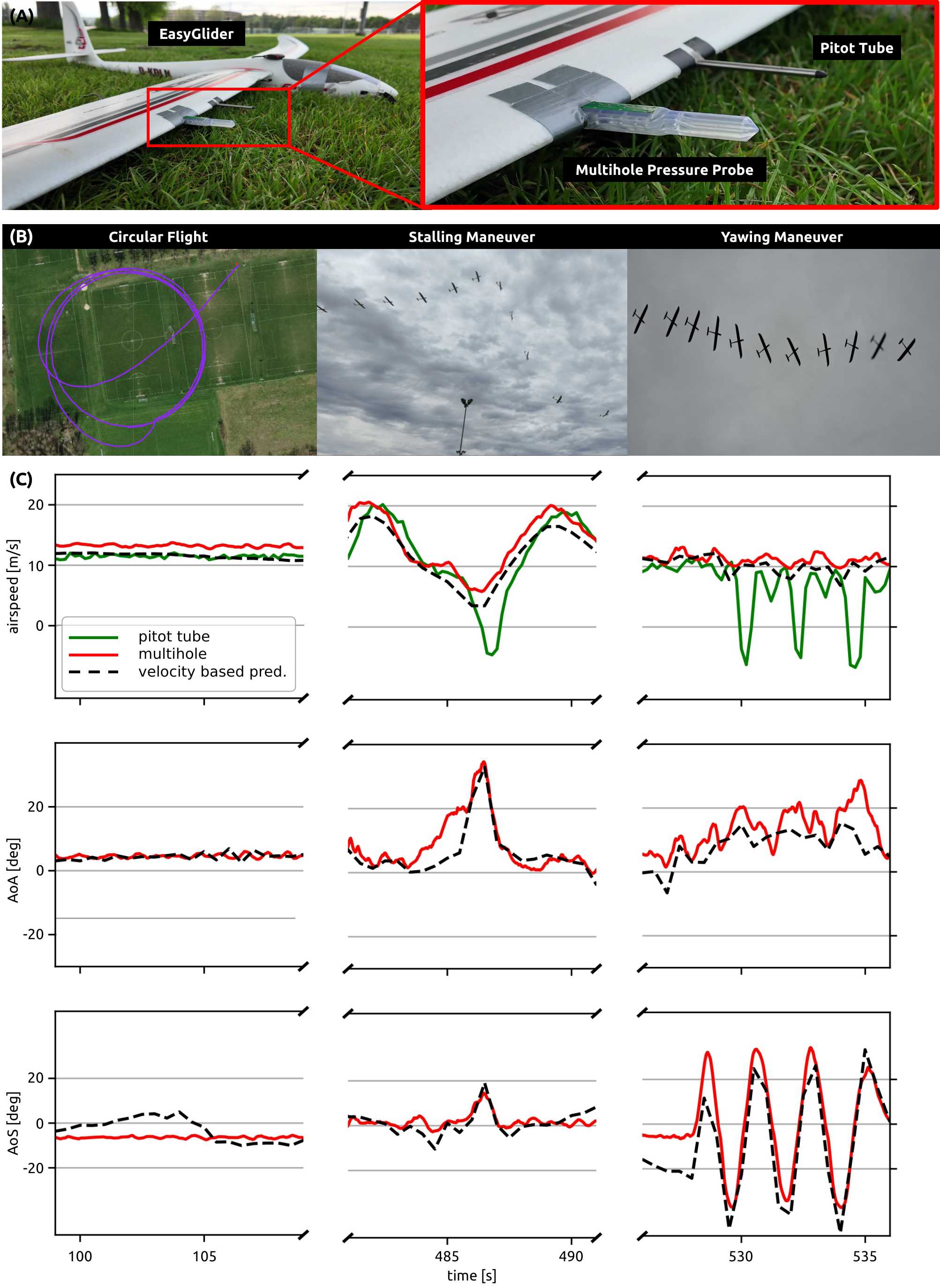}
    \caption{\textbf{Validation through outdoor flights}: \textbf{(A)} Experimental setup featuring an EasyGlider4 fixed-wing UAV equipped with a Pixhawk autopilot, a pitot tube and our MPP. \textbf{(B)} Three different manoeuvres were performed. \textbf{(C)} Estimations of airspeed, AoA, and AoS from our~MPP (red), the pitot tube (green) and velocity-based estimation from the Easyglider's autopilot (black).}
    \label{fig:validation}
\end{figure*}

\section{Outdoor Validation Flights} \label{sec:outdoor_validation}
Flight experiments were conducted with an EasyGlider4 fixed-wing UAV equipped with a Pixhawk autopilot. The MPP was mounted on the right wing adjacent to the pitot tube (Fig.~\ref{fig:validation}A). The flown flight manoeuvres consist of circular trajectories (constant altitude of 40~m, speed of 12~m/s), stalling and yawing manoeuvres (Fig.~\ref{fig:validation}B). \rev{Video material of the flights can be found here: \href{https://youtu.be/U3nR1v3fbZg}{https://youtu.be/U3nR1v3fbZg}.}

The Pixhawk autopilot performed state estimation of position, velocity, orientation, and airspeed using an IMU, GPS, and a pitot tube. From this data, the velocity in the body frame $\textbf{v} = [v_x, v_y, v_z]^T$ is estimated, allowing the calculation of the AoA $\alpha_{\text{est}}$ and AoS $\beta_{\text{est}}$.

\begin{align}
    \label{eq:px_est}
    \alpha_{\text{est}} &= tan^{-1}(v_z/v_x) \\
    \beta_{\text{est}} &= tan^{-1}(v_y/v_x) \nonumber
\end{align}

The estimates $v_x, \alpha_{\text{est}}$ and $\beta_{\text{est}}$ serve as a reference for comparison with the MPP; they do not, however, provide a ground truth. This comparison is only viable in the absence of wind, as the velocities estimated by the Pixhawk are in the body frame, while the pitot tube and the MPP provide estimates in the wind frame. Flights were conducted on days with no wind.

The comparison of the Pixhawk-generated estimation, the pitot tube-, and the MPP-measurements reveal three main observations (Fig.~\ref{fig:validation}C).
First, the MPPs airspeed measurements closely match the pitot tube during circular flights, achieving a MAE of \rev{1.65~m/s}. For stalling- and yawing-manoeuvres, the MPP outperforms the pitot tube, due to its ability to decouple airspeed and angle measurements.
Second, our AoA measurements closely match the autopilot's estimation, with a MAE of 3.20$^\circ$. This is important for early stall detection and prevention.
Finally, our AoS measurement agrees well with the autopilot's estimation during all manoeuvres and accurately captures yawing, with a MAE of 5.87$^\circ$.

This demonstrates that our sensor and its calibration perform well in an unseen, applied outdoor setting and did not overfit to the data collected in the controlled indoor experiments.

% \FloatBarrier
\section{Conclusion}
In this study, we developed and manufactured a nine-hole multihole pressure probe with varying tip shapes (spherical and conical) and hole spacings (0.4, 0.7, 0.9, and 1.2 mm). The probes were characterised in an open wind tunnel to identify the optimal design, focusing on maximising measurement resolution while minimising noise. The final design was calibrated using multivariate polynomial regression and validated through outdoor flight tests.
Weighing just 9 g and operating within an airspeed range of 3 to 27~m/s, our sensor addresses a critical niche for small UAVs. The airspeed, angle of attack, and angle of sideslip measurements demonstrated accurate flow estimation both in a controlled \rev{(MAE of 0.44~m/s, 0.163$^\circ$, and 0.156$^\circ$)} and outdoor setting \rev{(MAE of 1.65~m/s, 3.20$^\circ$, and 5.87$^\circ$)}.

\rev{A meaningful comparison with other multihole probes (Fig.~\ref{fig:state_of_the_art}) requires reporting both Root Mean Square Error (RMSE) and Mean Absolute Error (MAE) metrics, due to varying conventions across studies. Our probe achieves an RMSE of \(0.22^\circ\) for angle estimation and \(3.6\%\) for speed, positioning it between Witte~\cite{Witte_development_uav}~(\(<0.13^\circ\), \(<0.53\%\)) and Hall~\cite{Hall_oxford_probe}~(\(2^\circ\), \(4\%\)). In terms of MAE, our results (\(0.16^\circ\), \(2.9\%\)) outperform Rediniotis~\cite{Rediniotis_Tiny_probe} in angular accuracy (\(<0.28^\circ\)), while exhibiting a higher relative speed error (\(<0.35\%\)), likely due to their experiments being conducted at significantly higher airspeeds ($>$20 m/s).}

To expand our sensor's application range further, the calibration dataset could extend to higher angles ($>35^\circ$), enabling precise estimation during acrobatic manoeuvres. Additionally, the sensor's high measurement frequency of 50 Hz could be leveraged to analyse turbulent airflow in mid-air, enabling better disturbance rejection. These features could be particularly beneficial for highly agile UAVs, such as avian-inspired drones that utilise wing and tail morphing \cite{wuest2024agile,jeger2024adaptive,vu2024raptor,lentink2024tail} to maintain flight control at high angles of attack and in turbulent wind conditions.

\end{document}